%% file: root.tex
\documentclass[letterpaper, 10 pt, conference]{ieeeconf}  

\IEEEoverridecommandlockouts                              

\usepackage{graphics} 
\usepackage{epsfig} 
\usepackage{mathptmx} 
\usepackage{times} 
\usepackage{amsmath} 
\usepackage{amssymb}  
\usepackage{cite}
\usepackage{booktabs}
\usepackage{graphicx}
\usepackage{setspace}

\title{\LARGE \bf
Magnetic based In-situ Self 3D Pose Estimation for a Modular Soft Tendon-Driven Continuum Robot via IMU-Fusion
}

\author{Zheng Cao*, Guo Ning (Andrew) Sue*, Xiangyun Bu, David Quinn, Junzhe Hu, Carmel Majidi
\thanks{*Equal Contribution}
\thanks{All authors are from Carnegie Mellon University \tt\footnotesize{\{ 
zhengcao, gsue, xiangyun, djquinn, junzhehu, cmajidi\}@andrew.cmu.edu}}}

\begin{document}

\maketitle
\thispagestyle{empty}
\pagestyle{empty}

\begin{abstract}

Continuum robots are well suited for gentle manipulation because of their inherent compliance and ability to adapt to complex environments. However, their continuously deformable structure makes accurate configuration estimation challenging, particularly when external vision systems are unavailable or obstructed. In this work, we present an embedded pose sensing framework that combines inertial measurement units (IMUs) and active magnetic fields to estimate the robot configuration without relying on external cameras. The angular measurements from the IMU and magnetic-field references are fused to improve local orientation estimation and reduce accumulated orientation error during operation. This pose sensing scheme achieves an update rate of 16.7~Hz, allowing real-time feedback. The proposed system is experimentally validated through closed-loop control, where the estimated robot configuration is used to maintain the end-effector at a desired position while interacting with an object. These results demonstrate the potential of distributed magnetic--inertial sensing for real-time pose estimation and closed-loop control of continuum robots.

\end{abstract}

\section{INTRODUCTION}
\input{source/Intro}

\section{RELATED WORKS}
\input{source/RelatedWork}

\section{METHODOLOGY}
\input{source/Methodology}

\section{EXPERIMENTAL SETUP}
\input{source/ExperimentalSetup}

\section{RESULTS}
\input{source/Results}

\section{DISCUSSION}
\input{source/Discussion}

\section{CONCLUSIONS}
\label{sec:conclusion}
We presented a modular continuum robot combining PCB coils,
magnetometers, and onboard attitude sensing for shape estimation
without actuator models, strain sensors, or external cameras
during deployment. A shared learned joint model, per-segment
calibration, and quaternion--magnetic fusion achieved pooled
backbone and tip RMSEs of $10.0\pm3.2$ and
$14.9\pm6.5$~mm, respectively, across 20 trials in four
scenarios. Mean tip RMSE corresponded to 5.5\% of robot
length, with estimation demonstrated under both tendon
actuation and manual manipulation. The estimate also supported
tip-height control and disturbance recovery, although control
accuracy requires independent ground-truth validation.
Future work will improve drift correction, expand the training
workspace, extend feedback control from tip height to backbone
shape, and incorporate force sensing for more comprehensive
proprioception.

\addtolength{\textheight}{-12cm}   



\section*{ACKNOWLEDGMENT}

This work used large language models (LLMs) such as ChatGPT and Claude to assist in proofreading and revising the manuscript text.  In addition, LLMs were also used to assist in the creation of code segments/scripts used for experimental data analysis, figure generation and implementation.


\bibliographystyle{IEEEtran}
\bibliography{reference}

\end{document}

%% file: source/Intro.tex
Soft continuum robots deform continuously to conform to complex
environments and handle delicate objects, enabling gentle
manipulation~\cite{tang2026origami}, food
harvesting~\cite{li2026fruit}, and minimally invasive
procedures~\cite{lai2026single,yang2025mag, dupont2022continuum}. However, their
distributed deformation makes configuration estimation more
challenging than for rigid-link robots.

Reliable configuration feedback is essential for closed-loop
control. Estimates derived from actuation inputs through
kinematic or mechanical models~\cite{camarillo2009model,rao2021model}
are sensitive to material nonlinearities, hysteresis, friction,
and external loading~\cite{shen2026hys,mishra2022fractionalorder}.
External vision provides accurate measurements but requires
clear visibility~\cite{russo2023overview}, motivating embedded
proprioception. Strain sensors measure local deformation but
may require complex models for global shape
reconstruction~\cite{Liu2025sensor}. Accelerometers provide
inclination relative to gravity but cannot resolve full
orientation and are affected by dynamic
acceleration~\cite{martin2022proprioceptive}. Magnetic
measurements provide complementary directional information
for distributed shape estimation~\cite{sue2026tendon}.

\begin{figure}
    \centering
    \includegraphics[width=\linewidth]{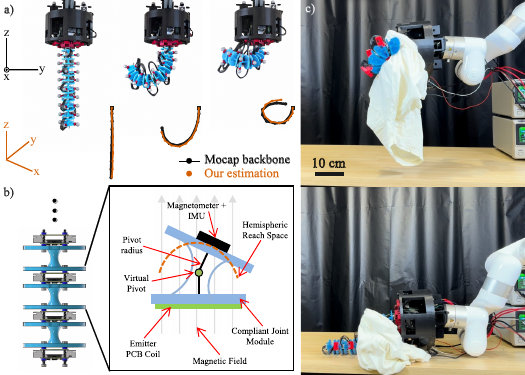}
    \caption{Modular tendon-driven continuum robot and sensing
principle. a) Pose estimation during 3D curling. Frontal
images show the physical robot, while isometric views
highlight the reconstructed three-dimensional backbone.
Orange denotes our estimate and black denotes motion-capture
ground truth. b) Modular platform design and sensing
principle. Relative motion between identical segments is
modeled using virtual pivots, with relative positions inferred
from magnetic fields generated by embedded PCB coils.
c) Grasping demonstration with a soft, deformable object
(T-shirt).}
    \label{fig:overview}
\end{figure}

Distributing magnetic--inertial sensing across modular robots
requires efficient acquisition and communication.
The approximately 1.2~Hz estimation rate
of~\cite{sue2026tendon} limits dynamic tracking and feedback
control. We address this limitation with embedded sensing
modules that combine onboard attitude estimates and magnetic
measurements to reconstruct the three-dimensional backbone.
Continuous attitude polling and intermittent magnetic
corrections enable configuration updates at 16.7~Hz.
The architecture supports distributed sensing and closed-loop
control under external loading without external visual tracking
during deployment.

The main contributions are:
\begin{enumerate}
    \item A modular magnetic--inertial sensing architecture
    for embedded three-dimensional shape reconstruction.
    \item A sensing and communication scheme providing
    configuration updates at 16.7~Hz.
    \item Experimental validation of shape estimation under
    varied deformation and contact conditions, and closed-loop
    tip-height control under an external disturbance.
\end{enumerate}

%% file: source/RelatedWork.tex
\subsection{Pose and Configuration Estimation}

Accurate configuration estimation is essential for continuum robots because their compliant, continuously deformable structures make the relationship between actuation and robot shape difficult to predict. Model-based approaches estimate robot configuration from actuation inputs using kinematic or mechanical models \cite{camarillo2009model,rao2021model}, but their accuracy can degrade in the presence of material nonlinearities, hysteresis, friction, and external loading.

External sensing has therefore been widely used to directly observe the robot configuration. Vision-based methods have employed stereo cameras and learning-based approaches for continuum-robot pose estimation \cite{reiter_learning_2011}, while other systems combine visual tracking with additional displacement measurements for real-time configuration estimation \cite{fang_design_2023}. Although these approaches can provide accurate measurements, they depend on external sensing infrastructure and an unobstructed line of sight.

Embedded deformation sensing provides an alternative for self-contained proprioception. Strain and fiber-optic sensing methods measure local deformation along the robot body and reconstruct the global configuration from distributed measurements \cite{lilge_continuum_2022,teetaert_stochastic_2025}. However, these approaches often require calibration or a model relating local sensor measurements to the global robot shape.

\subsection{Embedded Inertial and Magnetic Proprioception}

Inertial sensing has been explored as a compact approach for embedded configuration estimation.  \cite{hughes2021sensing} combined inertial sensing with actuator measurements to estimate continuum-robot pose, while \cite{martin2022proprioceptive} used distributed inertial measurement units (IMUs) to reconstruct robot shape and estimate the end-effector position. These studies demonstrate that distributed orientation measurements can provide proprioceptive information without external tracking.

Magnetic sensing has similarly been used for embedded shape estimation.  \cite{guo2019continuum} used magnetic measurements to infer local bending and reconstruct continuum-robot shape, while  \cite{baaij_learning_2023} and  \cite{adamu2025hall} demonstrated learning-based reconstruction from embedded magnetic-field measurements. Magnetic sensing provides an additional directional reference but can require dedicated magnetic sources, robot-specific calibration, or learned mappings.

More closely related to this work, magnetic and inertial measurements have been combined for embedded orientation estimation \cite{sue2026tendon}. However, existing implementations remain limited in update rate and scalability when distributed across long, modular continuum robots. In this work, we develop a modular magnetic--inertial proprioceptive sensing architecture that combines local orientation estimates from multiple robot segments to reconstruct the three-dimensional configuration at a rate suitable for real-time feedback control.

%% file: source/Methodology.tex
\label{sec:method}
\providecommand{\todo}[1]{\textbf{[TODO: #1]}}
\newcommand{\vecb}[1]{\mathbf{#1}}
\newcommand{\dB}{\Delta\vecb{B}}
\newcommand{\Tpair}{T_{\mathrm{pair}}}

\providecommand{\vecb}[1]{\mathbf{#1}}

\subsection{Design}
\begin{figure}
    \centering
    \includegraphics[width=\linewidth]{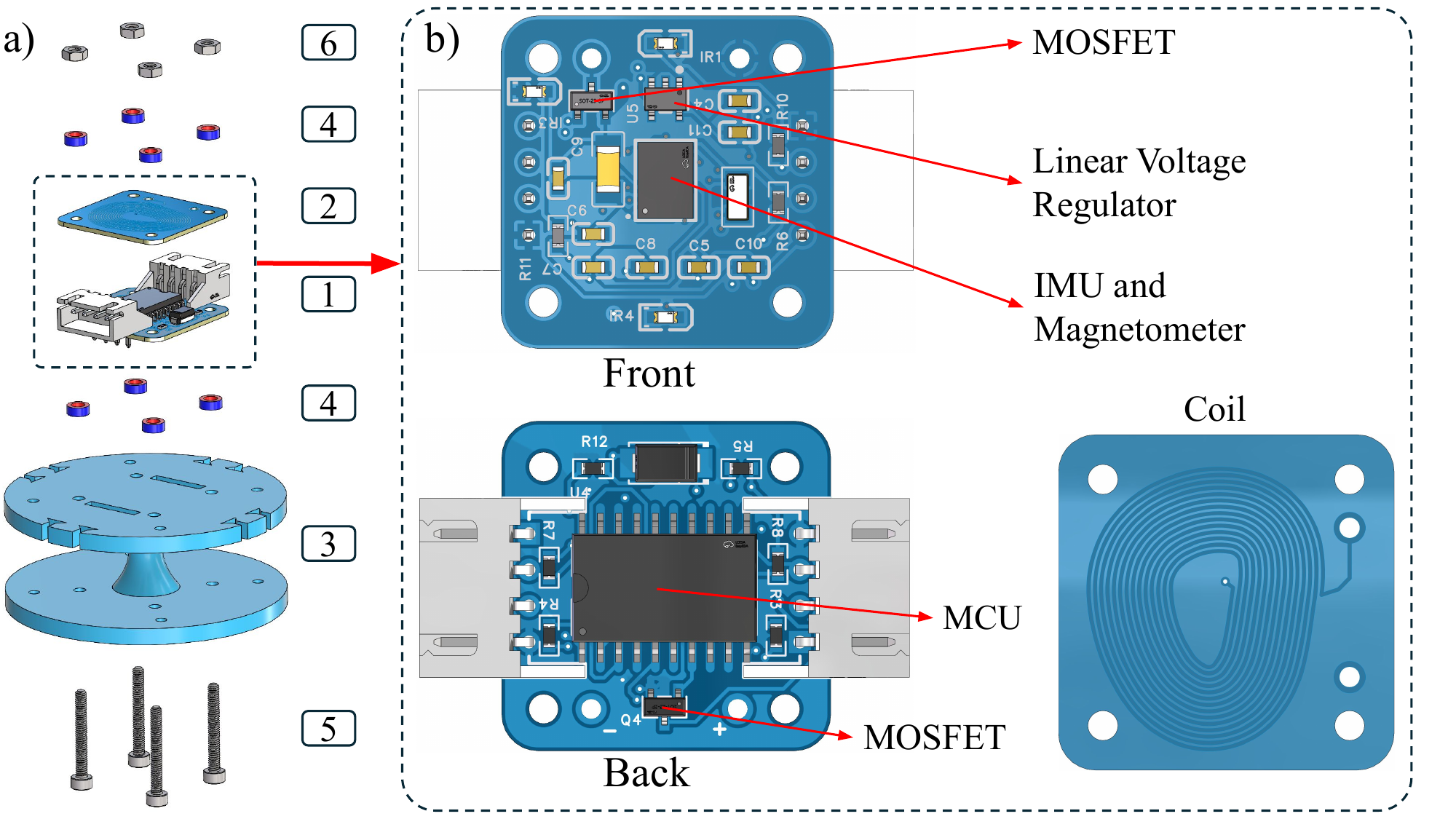}
    \caption{Overview of the mechanical and electronic components of one proprioceptive module of the continuum robot. a) Exploded view of a single module. Numbered parts include (1) PCB (2) coil (3) module (4) support standoff (5) M2-20mm screw (6) M2 nut. b) Customized sensing boards in each module and customized coil.}
    \label{fig:design}
\end{figure}
The robot comprises nine serial TPU 90A modules, each 20~mm
high, with two disks (50~mm in diameter, 2~mm thick) connected
by an hourglass joint (16~mm high, 5~mm minimum diameter). Compared to existing designs \cite{sue2026tendon} \cite{li2026hallimu}, our architecture features a soft mechanical module with integrated joints that simplifies the mechanical structure and reduces the number of separate components (Fig. \ref{fig:design} (a)). This approach facilitates assembly and disassembly while preserving modularity, allowing individual modules to be replaced and the robot length to be adjusted. In addition, each mechanical module has one magnetic coil instead of a single magnetic placed on the base, allowing infinite number of modules to be stacked. 
Each module carries a PCB integrating a BNO086 IMU,
ATtiny3226 microcontroller, planar coil, and switching
MOSFET (Fig. \ref{fig:design} (b)). Following~\cite{sue2026tendon}, each microcontroller
reads its IMU over SPI and exposes registers on a shared
I\textsuperscript{2}C bus. An ESP32-S3 polls the boards at
400~kHz and streams measurements over USB to a Raspberry
Pi~5 running ROS~2. Separate logic and coil rails isolate
the 3.3~V electronics from coil switching.

The BNO086 Game Rotation Vector fuses accelerometer and
gyroscope measurements without magnetic input, allowing
attitude tracking during coil activation. Magnetic
measurements correct its arbitrary initial heading and
subsequent drift. Raw magnetometer readings arrive at
approximately 90~Hz.

The wound coil in~\cite{sue2026tendon} is replaced by
additively wound traces on both sides of a 0.8~mm FR4 PCB
with 2~oz copper. The coil trace is defined by the following equation:
\begin{equation}
r(\phi)=R_0\!\left[1+\sum_{k=1}^{3}
\alpha_k\cos(k\phi+\beta_k)\right],
\label{eq:coil}
\end{equation}
where $\boldsymbol{\alpha}=(0.085,0.100,0.050)$ and
$\boldsymbol{\beta}=(0.5,0.2,-0.4)$~rad; successive turns
use constant pitch. Its asymmetric outline breaks axial
field symmetry (Fig. \ref{fig:design} (b)) and thus provide more features in the magnetic field to be captured and process down stream in the sensing pipelin. Approximately 19 turns per layer
of 10~mil traces give a resistance near $1\,\Omega$.
A 40~ms pulse at 3.3~V produces approximately
$100\,\mu$T at the neighboring segment, compared with
$0.6\,\mu$T noise per axis.

\subsection{Sensing}
\begin{figure*}
    \centering
    \includegraphics[width=\linewidth]{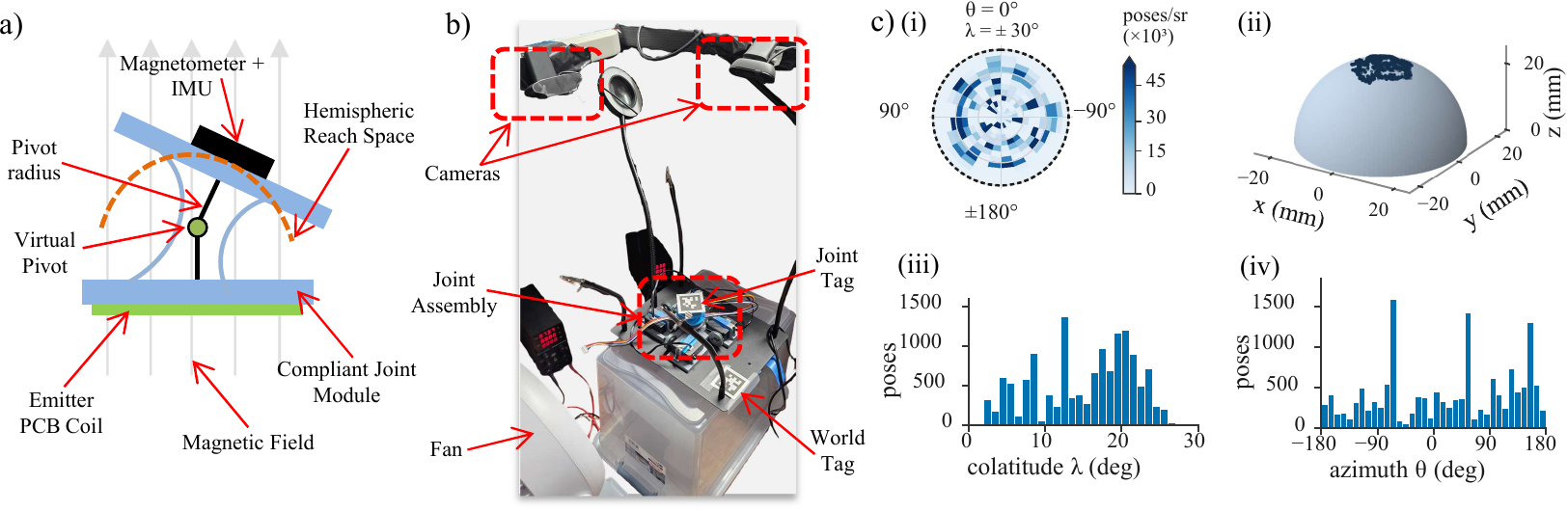}
    \caption{Sensing principle, data collection setup, and sampled pose
distribution. a) Joint bending and magnetic sensing under the fixed
pivot assumption of~\cite{sue2026tendon}. The IMU and magnetometer on
the upper segment are assumed to move along a hemispherical surface
at a fixed radius from a virtual pivot at the center of the compliant
joint. b) Data collection setup using two cameras to track a world
reference tag and a joint tag attached to the upper segment. The world
tag provides a common reference frame, allowing the joint position to
be measured when its tag is visible to either camera. Random target
orientations guide sampling across the reachable workspace to learn
the mapping from magnetic measurements to joint position.
c) Sampled pose distribution showing (i) positions across the
hemispherical workspace, (ii) a fitted spherical surface with an RMSE
of 0.19~mm, supporting the fixed pivot assumption, and (iii, iv)
histograms of colatitude and azimuth, respectively. Sampling covers
the workspace, although hardware biases and differences between
camera measurements may contribute to uneven coverage.}
    \label{fig:working_principles_distribution}
\end{figure*}
\subsubsection{Joint Model and Training}
Following~\cite{sue2026tendon}, each joint is modeled as
a fixed-radius spherical pivot without axial twist (Fig. \ref{fig:working_principles_distribution} (a)).
Its relative pose is therefore represented by a unit
direction $\vecb{u}\in S^2$. Unlike piecewise constant
curvature~\cite{pcc}, these local assumptions do not
prescribe bending along the full backbone. The pivot
radius is 10~mm; a 9~mm platform offset (height of sensing electronics), and a 10~mm before the pivot of the next module, giving a total pivot
spacing of $L=29$~mm.

Training used a fixed proximal coil and a distal sensor
actuated by four antagonistic tendons. Camera feedback
from two webcams and AprilTags guided the joint toward
targets sampled uniformly over a spherical cap (Fig. \ref{fig:working_principles_distribution} (b) (c)):
$\cos\lambda\sim\mathcal{U}[\cos\lambda_{\max},1]$ and
$\theta\sim\mathcal{U}[-\pi,\pi)$, with $\lambda_{\max}$
slightly below the $30^\circ$ mechanical limit. $\lambda$ and $\theta$ are colatitude and azimuth respectively. 
Equal-solid-angle Fibonacci cells and sampling quotas
improved coverage.

At each stationary target, two to three pulses, each
at most 40~ms, provided six coil-on samples. The coil
field was $\Delta\vecb{B}=\vecb{m}_{\mathrm{on}}-\vecb{b}(t)$,
where $\vecb{b}(t)$ was fitted linearly to surrounding
coil-off measurements (ambient). $\vecb{m}_{\mathrm{on}}$ denotes the measured magnetic reading when the coil was on. Approximately 15,000 unique poses
were collected over 10~h.

With the base leveled, normalized stationary
acceleration $\hat{\vecb{a}}=\vecb{a}/\|\vecb{a}\|$
provides the parent axis in the child sensor frame.
These training labels avoid camera and pivot-fitting
errors; cameras guide coverage and check consistency.
An MLP maps normalized $\Delta\vecb{B}$ to a unit vector,
using cosine loss
$\ell=1-f(\Delta\vecb{B})^\top\hat{\vecb{a}}$.
Hereafter, $f$ the function approximated by the MLP, includes input normalization, which
removes multiplicative field-amplitude variation.

\subsubsection{Acquisition}
Continuous attitude polling is interleaved with magnetic
bursts. Reading each 96-byte sensor block takes about
3.4~ms; a ten-segment (1 base module and 9 sensing module on TPU) sweep takes approximately 52~ms,
with an achieved attitude update rate of 16.7~Hz.
Every $T_{\mathrm{pair}}=300$~ms, one coil fires for
40~ms while the master polls only the other segments'
12-byte magnetometer blocks, taking approximately
3~ms per round.

For each receiver, the coil field is the mean of fresh
coil-on samples minus its latest confirmed coil-off
baseline. Samples are accepted only after settling and
a bitwise change indicating a new conversion. Each
segment receives a parent-coil correction every
$NT_{\mathrm{pair}}\approx3$~s, while attitude tracking
continues between bursts.

\subsubsection{Calibration and Reconstruction}
Let $R_i=R(\vecb{q}_i)R(\vecb{q}_i^0)^\top$, where
$\vecb{q}_i^0$ is recorded with the robot hanging
straight, and let $\vecb{e}=-\hat{\vecb{z}}$.
Quaternion and magnetic directions are
\begin{equation}
\begin{aligned}
\vecb{u}_i^q &= R_i\vecb{e},\\
\vecb{u}_i^m &= R_{i-1}M_{\vecb{e}}\,
f(A_iW_i\Delta\vecb{B}_i),\\
M_{\vecb{e}} &= 2\vecb{e}\vecb{e}^{\top}-I.
\end{aligned}
\label{eq:directions}
\end{equation}
Under the no-twist assumption, $M_{\vecb{e}}$ converts
the predicted parent axis in the child frame into the
child direction in the parent frame. The parent's
rotation is evaluated at coil activation.

Ambient subtraction removes constant offsets.
An ellipsoid fit to ambient-field measurements during
sensor rotation determines the gain and soft-iron
correction $W_i$. The remaining alignment $A_i$ is
fitted over a 45~s initialization window by minimizing
the angle between $M_{\vecb{e}}f(A_iW_i\Delta\vecb{B}_i)$
and $R_{i-1}^{\top}\vecb{u}_i^q$. Stationary data identify
tilt alignment; bending motion additionally identifies
axial alignment.

\subsubsection{Fusion}
Quaternion heading drift measured approximately
0.65\% of angular travel. We estimate its slowly varying
directional error using a per-segment rotation state
$\vecb{b}_i$:
\begin{equation}
\hat{\vecb{u}}_i
=\exp([\vecb{b}_i]_\times)\vecb{u}_i^q,
\qquad
P_i\leftarrow P_i+\sigma_b^2\Delta t\,I.
\label{eq:eskf-state}
\end{equation}
At each magnetic update,
\begin{equation}
\begin{aligned}
\vecb{y}_i &=
\operatorname{Log}(\vecb{u}_i^q\!\to\!\vecb{u}_i^m)
-\vecb{b}_i,\\
K_i &= P_i(P_i+\sigma_m^2I)^{-1},\\
\vecb{b}_i &\leftarrow \vecb{b}_i+K_i\vecb{y}_i,
\qquad P_i\leftarrow(I-K_i)P_i .
\end{aligned}
\label{eq:eskf-update}
\end{equation}
Here $\operatorname{Log}$ denotes the minimum rotation
between directions. The unobservable component about
$\vecb{u}_i^q$ is damped toward zero. The backbone is
reconstructed at the attitude update rate without
integrating acceleration:
\begin{equation}
\vecb{p}_0=\vecb{0},\qquad
\vecb{p}_i=\vecb{p}_{i-1}+L\hat{\vecb{u}}_i,
\quad i=1,\ldots,N.
\label{eq:backbone}
\end{equation}

\subsection{Closed-Loop Tip Height Control}
\label{sec:control}
The estimated tip height above the straight hanging
configuration is
\begin{equation}
h=L\sum_{i=1}^{N}(1+\hat{u}_{i,z}).
\label{eq:height}
\end{equation}
Height is independent of heading, so this task evaluates
height feedback rather than heading-drift correction.
A tendon is selected before each trial because bending
in either direction raises the tip. The spool velocity is
\begin{equation}
v=k_pe+k_iI_h-k_d\dot{h},\qquad
\dot{I}_h=e\,\mathbf{1}_{\{|e|<10\,\mathrm{mm}\}},
\quad e=h^*-h.
\label{eq:pid}
\end{equation}
The measured height derivative is low-pass filtered.
Conditional integration limits windup during slack
take-up; velocity magnitude and its rate of increase
are limited. Unwinding past neutral is prohibited
because it reverses the height response.

After alignment, the robot settles for 2~s at zero
command; median height and spool position establish
trial references. Trials are rejected if initial height
exceeds the straight configuration by 5~mm.

%% file: source/ExperimentalSetup.tex
\label{sec:setup}

\subsection{Platform and Ground Truth}
The robot comprised nine 29~mm segments
(261~mm total), suspended from a rigid base and
actuated by four antagonistic tendons using velocity-controlled
servomotors (Dynamixel XM540-W270-R) in the 4 cardinal directions. Unless stated otherwise, one tendon produced
unidirectional curling. The fused, quaternion, magnetic, and
inertial estimators processed identical recordings at 16.7~Hz
(Section~\ref{sec:method}). The ground truth from
the Optitrack Motive system recorded at
100~Hz was used. We report
position RMSE across all segments and frames, and tip RMSE.

\subsection{Pose Estimation Scenarios}
We evaluated four conditions:
\begin{enumerate}
    \item \textbf{Planar curl (2D curl):} Slow, unobstructed single-tendon
    bending constrained to a plane on a flat surface.
    \item \textbf{Planar curl with obstacle (2D obst.):} The same motion
    against a box of rocks, whose location was unknown to
    the estimator, testing contact-induced deformation.
    \item \textbf{Free curl in 3D (3D curl):} Single-tendon actuation
    while freely suspended, testing spatial reconstruction
    and magnetic correction of quaternion heading error.
    \item \textbf{Manual manipulation in 3D (3D hand):} Arbitrary
    bending with slack tendons, including shapes beyond
    the actuated workspace, testing externally imposed
    deformation without actuator or tendon-length inputs.
\end{enumerate}

\subsection{Demonstrations}
\subsubsection{Cloth Grasping}
Mounted on a UFactory Xarm7, the robot curled around
and lifted a shirt without feedback control. We present
the estimated backbone during the task, demonstrating
onboard sensing during occlusion by the cloth and arm.

\subsubsection{Closed-Loop Height Hold}
The controller in Section~\ref{sec:control} used fused
feedback to maintain the tip 40~mm above its relaxed
vertical position. Height,
$h=L\sum_i(1+u_{i,z})$, is independent of heading error.
We compare target, estimated, and independently measured
heights, and assess disturbance recovery after attaching
a 500~g tip load. Furthermore, to demonstrate the modular aspect of our paradigm, we used only 8 modules (9 including the base) for this experiment where as every other experiments we used 9 modules (10 , including the base)

\subsection{Drift and Hysteresis}
\subsubsection{Cyclic Actuation}
One tendon underwent 50 pull--release cycles of half a
spool turn, each lasting 10~s (5~s per direction;
500~s total). Encoder feedback at 50~Hz tracked
$p^*(t)=p_0+\tfrac{A}{2}[1-\cos(2\pi t/T)]$,
ensuring identical commanded strokes and return positions.
Pulling and releasing estimates were compared at matched
spool positions to assess hysteresis, repeatability and drift.

\subsubsection{Static Hold}
The robot remained in a fixed curled configuration for
30~min. Joint-angle and tip-position changes quantified
estimator drift, particularly for the quaternion chain
and fused estimate.

%% file: source/Results.tex
\label{sec:results}

\begin{figure*}[t]
\centering
\includegraphics[width=\linewidth]{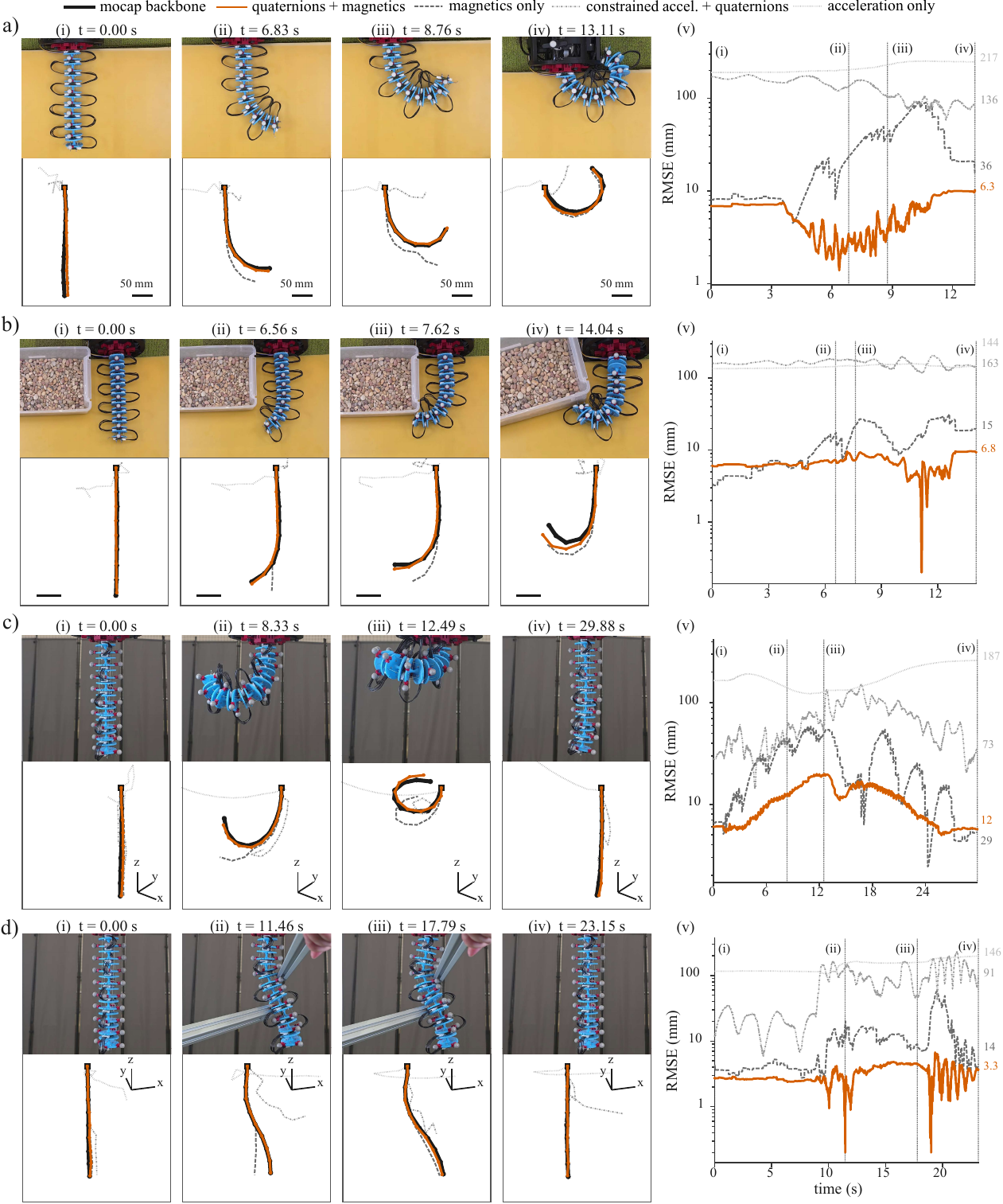}
\caption{Pose estimation during a) planar curling,
b) obstacle contact, c) free 3D curling, and
d) manual manipulation. Panels (i) to (iv) show selected
robot configurations and the corresponding estimated backbones.
Panel (v) shows instantaneous backbone RMSE against motion
capture, with errors at the final frame annotated.
The fused estimate (our estimation) is shown in orange, the magnetic estimate
as a dashed line, constrained acceleration with quaternions
as a dash dotted line, and acceleration alone as a dotted line.
Motion capture is shown in solid black.}
\label{fig:results}
\end{figure*}

\begin{table*}[t]
\centering
\caption{Position estimation accuracy and reported drift.
Scenario RMSEs are mean $\pm$ standard deviation against
motion capture over five trials; Avg.\ reports the mean
$\pm$ standard deviation across all 20 trials pooled across
the four scenarios, computed from the rounded scenario statistics.
Drift values are changes per 50~s averaged across
$x$, $y$, and $z$ during recordings without ground truth
and may include physical motion.}
\label{tab:rmse}
\resizebox{\textwidth}{!}{%
\begin{tabular}{@{}lrrrrrrr@{}}
\toprule
& \multicolumn{5}{c}{Backbone RMSE (mm)}
& \multicolumn{2}{c}{Drift (mm/50\,s)} \\
\cmidrule(lr){2-6} \cmidrule(l){7-8}
Method & 2D curl & 2D obst. & 3D curl & 3D hand
& Avg. & Cyclic & Hold \\
\midrule
Fusion (this work)
& $7.6\pm1.1$ & $11.3\pm1.1$ & $13.3\pm0.9$
& $7.8\pm4.0$ & $10.0\pm3.2$ & 4.1 & 1.2 \\
Magnetics only
& $36.9\pm2.2$ & $18.0\pm1.5$ & $29.4\pm0.9$
& $14.1\pm2.7$ & $24.6\pm9.5$ & 51.1 & 3.3 \\
Constrained acceleration + quaternions
& $133.8\pm3.4$ & $163.0\pm3.4$ & $71.9\pm2.1$
& $90.5\pm4.3$ & $114.8\pm36.8$ & 54.6 & 17.1 \\
Acceleration only
& $189.0\pm20.4$ & $153.8\pm15.3$ & $177.7\pm9.1$
& $180.8\pm21.1$ & $175.3\pm20.7$ & 24.5 & 0.3 \\
\midrule
& \multicolumn{5}{c}{Tip RMSE (mm)} & & \\
\cmidrule(lr){2-6}
Method & 2D curl & 2D obst. & 3D curl & 3D hand
& Avg. & & \\
\midrule
Fusion (this work)
& $8.8\pm1.1$ & $16.5\pm1.9$ & $21.1\pm1.8$
& $13.0\pm9.5$ & $14.9\pm6.5$ & & \\
Magnetics only
& $59.9\pm2.6$ & $36.2\pm3.0$ & $48.0\pm1.4$
& $24.9\pm5.9$ & $42.3\pm13.8$ & & \\
Constrained acceleration + quaternions
& $193.7\pm9.2$ & $246.0\pm5.4$ & $113.1\pm6.8$
& $151.8\pm7.8$ & $176.2\pm51.1$ & & \\
Acceleration only
& $297.5\pm35.1$ & $232.6\pm32.1$ & $279.4\pm12.9$
& $289.1\pm32.8$ & $274.7\pm37.4$ & & \\
\bottomrule
\end{tabular}%
}
\end{table*}

\subsection{Pose Estimation Against Motion Capture}
Across four scenarios and 20 trials, fusion achieved
scenario-averaged backbone and tip RMSEs of 10.0 and
14.9~mm, respectively, with tip error equal to 5.5\%
of the 270~mm robot length. Link angular RMSE was
$9.3^\circ$ (Table~\ref{tab:rmse},
Fig.~\ref{fig:results}).

Magnetic-only backbone and tip RMSEs averaged 24.6
and 42.3~mm, partly reflecting sparse updates during
motion. At coil activation, magnetic angular errors
were $5.4^\circ$ median, $9.7^\circ$ RMS, and
$17.8^\circ$ at the 95th percentile.
The constrained and unconstrained inertial variants
saturated in every trial, with backbone RMSEs of
114.8 and 175.3~mm, consistent with joint accelerations
below residual accelerometer bias.
Position errors generally accumulated toward the tip.

\subsubsection{Scenario Comparison}
Planar curling produced fused backbone and tip RMSEs
of 7.6 and 8.8~mm, with $6.3^\circ$ angular RMSE.
Obstacle contact increased these to 11.3 and 16.5~mm
and $9.7^\circ$. Compression violated the fixed-length
approximation, while marker reconstruction in
10--14\% of frames contributed to an 8.8~mm metric
floor, compared with 2.0~mm without contact.

Free 3D curling produced the largest errors:
13.3~mm backbone and 21.1~mm tip RMSE.
Joint bends reached $20^\circ$--$40^\circ$, exceeding
the $25^\circ$--$30^\circ$ training cap.
Magnetic angular errors reached a $7.4^\circ$ median
and $31.7^\circ$ 95th percentile; errors decreased
as the robot straightened.

Manual manipulation achieved 7.8~mm backbone and
13.0~mm tip RMSE despite occlusion, twisting, and
compression, without actuator inputs. The best trial
achieved 3.5~mm backbone RMSE against a 1.4~mm
metric floor.

\subsubsection{Model Variants and Alignment}
Five training and architecture variants differed
by less than 0.4~mm in fused backbone RMSE and
$0.2^\circ$ in magnetic angular error.
Receiver alignment had a larger effect, with its
omission increasing error for both fused and
magnetic-only estimation.

\subsection{Demonstrations}
\subsubsection{Cloth Grasping}
Estimation continued during occlusion by the cloth
and arm (Fig.~\ref{fig:demos_and_drift}(a)).
Across stationary intervals lasting 7, 20, and 14~s,
tip estimates varied by at most 6~mm for fusion,
40~mm for magnetic-only estimation, and 50--280~mm
for the inertial variants. These measurements
characterize stability rather than absolute accuracy.

\subsubsection{Closed-Loop Height Hold}
Fused feedback at 18.6~Hz, using 8 modules, regulated tip height to
40~mm (Fig.~\ref{fig:demos_and_drift}(b)).  
The reconstructed response reached the target in
3.4~s, peaked at 46~mm (15\% overshoot), and settled
within $\pm2$~mm after 5.5~s, subsequently holding
$40.6\pm0.04$~mm for 10~s.

A 500~g load applied at 24~s lowered the tip and
redistributed bending from distal to proximal joints.
Height recovered to within 2~mm of the target
5.1~s after its minimum and subsequently held
$39.0\pm0.3$~mm.
Furthermore, in this scenario we showed that our paradigm work just as well with one less module that the other scenarios, demonstrating our modular ability.

\subsection{Long-Term Estimation}
\subsubsection{Cyclic Actuation}
Across 49 analyzed cycles over 500~s, the final
phase-matched tip difference reached 29.1~mm for
fusion; magnetic-only differences ranged from
29 to 44~mm without a clear trend
(Fig.~\ref{fig:demos_and_drift}(c)(ii)).
Baseline change and cycle-to-cycle variation were
16.1 and 2.3~mm for fusion, versus 4.4 and 7.9~mm
for magnetic-only estimation. Magnetic measurements
thus showed less baseline change but greater
variability. Without ground truth, these changes
cannot separate sensor drift from creep or imperfect
motion repetition.

\subsubsection{Static Hold}
During a 31.9~min interval with all gyroscopes below
0.02~rad/s, 5258 magnetic fixes were recorded.
Fused and magnetic-only tip displacements reached
11.1 and 12.7~mm, respectively, with linear-trend
displacements of 8.7 and 8.4~mm and RMS scatter of
0.6 and 1.3~mm
(Fig.~\ref{fig:demos_and_drift}(c)(i)).
The shared downward trend was consistent with curl
relaxation, but does not establish an absolute drift
bound without independent ground truth.

%% file: source/Discussion.tex
\label{sec:discussion}

\begin{figure*}[t]
\centering
\includegraphics[width=\linewidth]{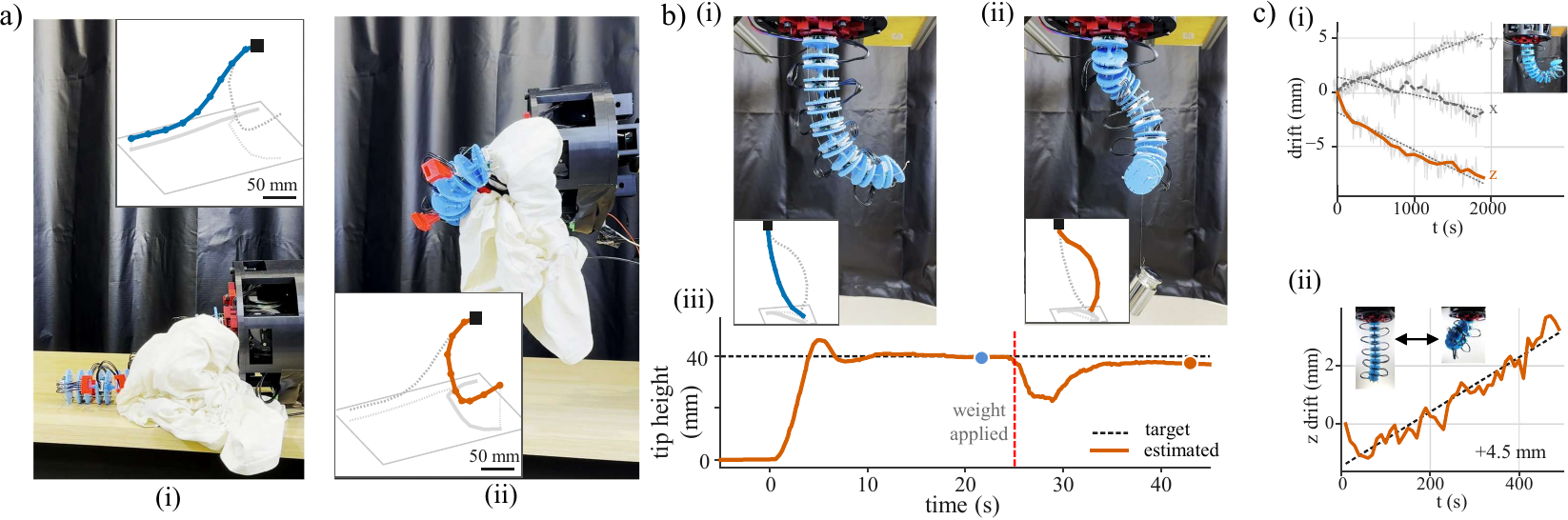}
\caption{Task demonstrations and long-term estimation.
a) Open-loop shirt grasping with grip attachments:
(i) initial and (ii) final configurations.
b) Height regulation at 40~mm under a 500~g tip load:
(i--ii) configurations before and after loading;
(iii) height response. Blue and orange dots mark
the illustrated frames; the red dashed line marks
load application. For this demonstration, we removed the end module to demonstrate that the shape estimation ability is independent of the number of modules. 
c) Estimated displacement during (i) a stationary
hold and (ii) cyclic actuation. The cyclic plot shows
$z$ displacement relative to the first cycle.}
\label{fig:demos_and_drift}
\end{figure*}

\begin{table*}[!t]
\centering
\footnotesize
\caption{Reported proprioceptive sensing accuracy.
$L$: robot or sensed length; ---: unreported.
Metrics and experimental conditions differ across studies.
For our work, the row averages scenario means; this work's
shape error is backbone RMSE across nine tracked segments.}
\label{tab:shape_sensing_comparison}
\setlength{\tabcolsep}{4pt}
\renewcommand{\arraystretch}{1.05}
\begin{tabular}{@{}lrlll@{}}
\toprule
Work & $L$ [mm] & Sensing / method & Tip error & Shape error \\
\midrule
Baaij et al.~\cite{baaij_learning_2023}
& 110 & 3 magnetometers + magnet, NN
& --- & 4.5\% (config.) \\
Adamu et al.~\cite{adamu2025hall}
& 240 & Hall + magnet/segment, MLP
& $<$5\%$L$ & max 11\,mm per marker \\
Guo et al.~\cite{guo2019continuum}
& --- & embedded magnets + sensors
& 0.23--9.34\,mm & --- \\
Li et al.~\cite{li2026shape}
& --- & Hall + base magnet + IMUs
& --- & $<$1\,mm \\
Pittiglio et al.~\cite{pittiglio2024magnetic}
& 64 & ball chain + Hall array
& 2.9\%$L$ (max 7.1\%) & --- \\
Thuruthel et al.~\cite{thuruthel2019soft}
& 120 & cPDMS strain, LSTM
& $2.4\pm2.2$\,mm & --- \\
Vicari et al.~\cite{vicari2023proprioceptive}
& 220 & base pressure, BiLSTM
& $6.0\pm4.7$\,mm (2.7\%$L$) & --- \\
Martin et al.~\cite{martin2022proprioceptive}
& 660 & 8 IMUs, PCC
& median $<$10\%$L$ & --- \\
Stella et al.~\cite{stella_soft_2024}
& 750 & 2 IMUs, PCC filter
& $\approx$7\%$L$ & --- \\
Han et al.~\cite{han2025enhanced}
& 583 & bend + IMUs, Kalman
& RMSE 17.0\,mm (2.9\%$L$) & --- \\
Sue et al.~\cite{sue2026tendon}
& 650 & coils + magnetometers, NN
& 47.4\,mm (7.3\%$L$) & RMSE 44.2\,mm \\
\textbf{This work}
& $\approx$270 & PCB coils + IMUs, NN + ESKF
& 14.9\,mm (5.5\%$L$) & RMSE 10.0\,mm \\
\bottomrule
\end{tabular}
\end{table*}
\subsection{Effect of Magnetic Fusion}
Magnetic corrections offered little benefit during
13--30~s trials. Magnetic estimates had $5.4^\circ$
median angular error and updated every 3~s per joint,
whereas attitudes updated at 16.7~Hz with heading drift
of approximately 0.65\% of angular travel. Greater
magnetic weighting may therefore degrade short-term
accuracy while potentially helping after larger
accumulated motion.

Across 49 cycles, fused and magnetic-only baseline
changes were approximately 16 and 4~mm, with
cycle-to-cycle variations of 2.3 and 7.9~mm.
Filter parameters came from holdout residuals and
measured heading drift without further tuning.
Retuning or joint-specific bias modeling may improve
correction at the cost of greater variability.
Both estimators showed similar 8--9~mm trends during
the 32~min static hold, consistent with creep but
not excluding shared estimation error.

Fusion reconstructed the backbone without acceleration
integration, avoiding the 100--300~mm errors developed
by inertial variants within 1~s. Fused feedback also
supported height regulation and disturbance recovery,
with a reconstructed settling time of 5.5~s to
$\pm2$~mm. However, height is heading-insensitive,
and the response was reconstructed from logged
attitudes because the controller's estimate was
not recorded.

Overall tip RMSE was 14.5~mm (5.4\% of length),
with 3.3\% during planar curling and 5--8\% in 3D.
These values are comparable to reported small magnetic
systems and below the approximately 7--10\% reported
for longer IMU-only arms, but exceed the 2.9\% reported
for bend-sensor--IMU fusion
(Table~\ref{tab:shape_sensing_comparison}).
Our system uses no strain or bend sensors and shares
one trained joint model, although differing platforms
and metrics limit direct comparison.

\subsection{Measurement and Evaluation Limitations}
RMSE includes reference, reconstruction, and alignment
uncertainty. Planar references used single markers
with possible center offsets; 3D references used three
markers, inferring missing rotation from chain pitch
when only two were visible. 

Obstacle contact compressed joints and required
reconstruction of interior segments in 10--14\% of
frames. Restricting evaluation to measured positions
reduced fused RMSE from 11.3 to 10.5~mm, indicating
that reference limitations contributed to contact error.
Friction, actuation speed, marker occlusion, and
manual placement introduced further variation.
Tight curls exceeded the training range, while manual
forces could violate the no-twist and fixed-length
assumptions. Hardware and calibration also varied
between sessions.

A shared time offset and world heading were fitted
from segment attitudes. Independent magnetic alignment
reduced magnetic RMSE from 25.2 to 15.1~mm, while
fused RMSE changed by only 0.1~mm, demonstrating
sensitivity to the comparison frame. Receiver alignment
also used attitudes from other trials in the same
scenario, excluding the scored trial and motion capture.
The sensing channels are therefore not fully independent,
and a single world heading cannot correct relative
segment drift.

Without independent ground truth, long recordings
cannot separate creep from estimator drift, and
height-control accuracy remains unverified.
These limitations are insufficient to explain the
much larger inertial errors but constrain conclusions
about differences of a few millimetres between methods.